\documentclass[runningheads]{llncs}

\usepackage{graphicx}
\usepackage{amsmath}
\usepackage{amsfonts}
\usepackage{bm}
\usepackage[misc]{ifsym}
\begin{document}
	\title{GIKT: A Graph-based Interaction Model for Knowledge Tracing}
	
	
	
	\author{Yang Yang\inst{1} \and
		Jian Shen\inst{1} \and
		Yanru Qu\inst{2}\and Yunfei Liu\inst{1} \and Kerong Wang\inst{1}\and Yaoming Zhu\inst{1}\and Weinan Zhang\inst{1}(\Letter) \and Yong Yu\inst{1}(\Letter)}
	\authorrunning{Y. Yang et al.}
	
	\institute{Shanghai Jiao Tong University
		\\ 
		\email{$\{$yyang,rockyshen,ymzhu,yyu$\}$@apex.sjtu.edu.cn,\\
			$\{$liuyunfei,wangkerong,wnzhang$\}$@sjtu.edu.cn}\and
		University of Illinois, Urbana-Champaign	\\\email{yanruqu2@illinois.edu}}
	
	\maketitle              
\begin{abstract}
	With the rapid development in online education, \textit{knowledge tracing} (KT) has become a fundamental problem which traces students' knowledge status and predicts their performance on new questions. Questions are often numerous in online education systems, and are always associated with much fewer skills. However, the previous literature fails to involve question information together with high-order question-skill correlations, which is mostly limited by data sparsity and multi-skill problems. From the model perspective, previous models can hardly capture the long-term dependency of student exercise history, and cannot model the interactions between student-questions, and student-skills in a consistent way. In this paper, we propose a Graph-based Interaction model for Knowledge Tracing (GIKT) to tackle the above probems. More specifically, GIKT utilizes graph convolutional network (GCN) to substantially incorporate question-skill correlations via embedding propagation. Besides, considering that relevant questions are usually scattered throughout the exercise history, and that question and skill are just different instantiations of knowledge, GIKT generalizes the degree of students' master of the question to the interactions between the student's current state, the student's history related exercises, the target question, and related skills. Experiments on three datasets demonstrate that GIKT achieves the new state-of-the-art performance, with at least 1\% absolute AUC improvement.

\end{abstract}
\section{Introduction}

In online learning platforms such as MOOCs or intelligent tutoring systems, \textit{knowledge tracing} (KT)~\cite{corbett1994knowledge} is an essential task, which aims at tracing the knowledge state of students. At a colloquial level, KT solves the problem of predicting whether the students can answer the new question correctly according to their previous learning history. The KT task has been widely studied and various methods have been proposed to handle it.

\begin{figure}
	\centering
	\includegraphics[width=0.4\textwidth]{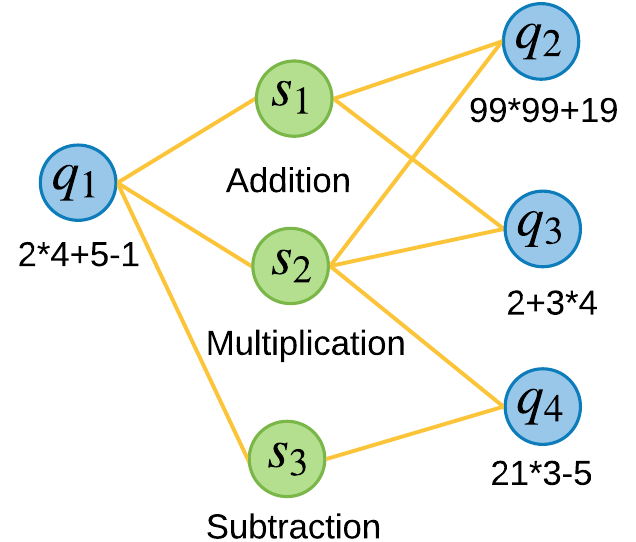}
	\caption{A simple example of question-skill relation graph. } \label{example}
\end{figure}

Existing KT methods \cite{dkt,dkvmn,d2008more} commonly build predictive models based on the skills that the target questions correspond to rather than the questions themselves. In the KT task, there exists several skills and lots of questions where one skill is related to many questions and one question may correspond to more than one skill, which can be represented by a relation graph such as the example shown in Figure \ref{example}. Due to the assumption that skill mastery can reflect whether the students are able to answer the related questions correctly to some extent, it is a feasible alternative to make predictions based on the skills just like previous KT works.

Although these pure skill-based KT methods have achieved empirical success, the characteristics of questions are neglected, which may lead to degraded performance. For instance, in Figure \ref{example}, even though the two questions $q_2$ and $q_3$ share the same skills, their different difficulties may result in different probabilities of being answered correctly. To this end, several previous works \cite{minn2019dynamic} utilize the question characteristics as a supplement to the skill inputs. 
However, as the number of questions is usually large while many students only attempt on a small subset of questions, most questions are only answered by a few students, leading to the data sparsity problem \cite{dhkt}. Besides, for those questions sharing part of common skills 
(\textit{e.g.} $q_1$ and $q_4$), simply augmenting the question characteristics loses latent inter-question and inter-skill information. Based on these considerations, it is important to exploit high-order information between the questions and skills.

In this paper, we first investigate how to effectively extract the high-order relation information contained in the question-skill relation graph. Motivated by the great power of Graph Neural Networks (GNNs) \cite{gat,kipf2016semi,hamilton2017inductive} to extract graph representations by aggregating information from neighbors, we leverage a graph convolutional network (GCN) to learn embeddings for questions and skills from high-order relations. Once the question and skill embeddings are aggregated, we can directly feed question embeddings together with corresponding answer embeddings as the input of KT models. 

In addition to the input features, another key issue in KT is the model framework. Recent advances in deep learning simulate a fruitful line of deep KT works, which leverage deep neural networks to sequentially capture the changes of students' knowledge state. Two representive deep KT models are Deep Knowledge Tracing (DKT) \cite{dkt} and Dynamic Key-Value Memory Networks (DKVMN) \cite{dkvmn} which leverage Recurrent Neural Networks (RNN) \cite{williams1989learning} and Memory-Augmented Neural Networks (MANN) respectively to solve KT. However, they are notoriously unable to capture long-term dependencies in a question sequence \cite{skvmn}. To handle this problem, Sequential Key-Value Memory Networks (SKVMN) \cite{skvmn} proposes a hop-LSTM architecture that aggregates hidden states of similar exercises into a new state and Exercise-Enhanced Recurrent Neural Network with Attention mechanism (EERNNA) \cite{su2018exercise} uses the attention mechanism to perform weighted sum aggregation for all history states.

Instead of aggregating related history information into a new state for prediction directly, we take a step further towards improving long-term dependency capture and better modeling student's mastery degree.
Inspired by SKVMN and EERNNA, we introduce a recap module to select several the most related hidden exercises according to the attention weight with the intention of noise reduction. Considering the mastery of the new question and its related skills, we generalize the interaction module and interact the relevant exercises and the current hidden states with the aggregated question embeddings and skill embeddings. The generalized interaction module can better model student's mastery degree of question and skills. Besides, an attention mechanism is applied on each interaction to make final predictions, which automatically weights the prediction utility of all the interactions.  



To sum up, in this paper, we propose an end-to-end deep framework, namely Graph-based Interaction for Knowledge Tracing (GIKT), for knowledge tracing. Our main contributions are summarized as follows: 1) By leveraging a graph convolutional network to aggregate question embeddings and skill embeddings, GIKT is capable to exploit high-order question-skill relations, which mitigates the data sparsity problem and the multi-skill issue. 2) By introducing a recap module followed by an interaction module, our model can better model the student's mastery degree of the new question and its related skills in a consistent way.
3) Empirically we conduct extensive experiments on three benchmark datasets and the results demonstrate that our GIKT outperforms the state-of-the-art baselines substantially.




\section{Related Work}

\subsection{Knowledge Tracing}

Existing knowledge tracing methods can be roughly categorized into two groups: traditional machine learning methods and deep learning methods. In this paper, we mainly focus on the deep KT methods.

Traditional machine learning KT methods mainly involve two types: Bayesian Knowledge Tracing (BKT) \cite{corbett1994knowledge} and factor analysis models. BKT is a hidden Markov model which regards each skill as a binary variable and uses bayes rule to update state. Several works extends the vanilla BKT model to incorporate more information into it such as slip and guess probabilty \cite{d2008more}, skill difficulty \cite{pardos2011kt} and student individualization \cite{pardos2010modeling,yudelson2013individualized}. 
On the other hand, factor analysis models focus on learning general paramaters from historical data to make predictions. Among the factor analysis models, Item Response Theory (IRT) \cite{ebbinghaus2013memory} models parameters for student ability and question difficulty, Performance Factors Analysis (PFA)         \cite{pavlik2009performance} takes into account the number of positive and negative responses for skills  and Knowledge Tracing Machines \cite{vie2019knowledge} leverages Factorization Machines \cite{rendle2010factorization} to encode side information of questions and users into the parameter model. 

Recently, due to the great capacity and effective representation learning, deep neural networks have been leveraged in the KT literature. Deep Knowledge Tracing (DKT) \cite{dkt} is the first deep KT method, which uses recurrent neural network (RNN) to trace the knowledge state of the student. Dynamic Key-Value Memory Networks (DKVMN) \cite{dkvmn} can discover the underlying concepts of each skill and trace states for each concept. Based on these two models, several methods have been proposed by considering more information, such as the forgetting behavior of students \cite{nagatani2019augmenting}, multi-skill information and  prerequisite skill relation graph labeled by experts \cite{chen2018prerequisite} or student individualization \cite{minn2018deep}. GKT \cite{nakagawa2019graph} builds a skill relation graph and learns their relation explicitly. However, these methods only use skills as the input, which causes information loss.

Some deep KT methods take question characteristices into account for predictions. Dynamic Student Classification on Memory Networks (DSCMN) \cite{minn2019dynamic} utilizes question difficulty to help distinguish the questions related to the same skills. Exercise-Enhanced Recurrent Neural Network with Attention mechanism (EERNNA) \cite{su2018exercise} encodes question embeddings using the content of questions so that the question embeddings can contain the characteristic information of questions, however in reality it is difficult to collect the content of questions. Due to the data sparsity problem, DHKT \cite{DBLP:conf/edm/WangMG19} augments DKT by using the relations between questions and skills to get question representations, which, however, fails to capture the inter-question and inter-skill relations. In this paper we use GCN to extract the high-order information contained in the question-skill graph. To handle the long term dependency issue, Sequential Key-Value Memory Networks (SKVMN) \cite{skvmn} uses a modified LSTM with hops to enhance the capacity of capturing long-term dependencies in an exercise sequence. And EERNNA \cite{su2018exercise} assumes that current student knowledge state is a weighted sum aggregation of all historical student states based on correlations between current question and historical quesitons. Our method differs from these two works in the way that they aggregate related hidden states into a new state for prediction, while we first select the most useful history exercises to reduce the effects of the noisy in the current state, and then we perform pairwise interaction for prediction. 

\subsection{Graph Neural Networks}


In recent years, graph data is widely used in deep learning models. However, the traditional neural network suffers from the complex non-Euclidean structure of graph. 
Inspired by CNNs, some works use the convolutional method for the graph-structure data \cite{kipf2016semi,defferrard2016convolutional}. Graph convolutional networks (GCNs) \cite{kipf2016semi} is proposed for semi-supervised graph classification, which updates node representations based on itself and its neighbors. In this way, updated node representations contain attributes of neighbor nodes and information of high-order neighbors  if multiple graph-convolutional layers are used. Due to the great success of GCNs, some variants are further proposed for graph data \cite{gat,hamilton2017inductive}.

With the development of Graph Neural Networks (GNNs), many applications based on GNNs appear in various domains, such as natural language processing (NLP) \cite{beck2018graph,yao2019graph}, computer vision (CV) \cite{qi20173d,garcia2017few} and recommendation systems \cite{wang2019kgat,qu2019end}. As GNNs help to capture high-order information, we use GCN in our GIKT model to extract relations between skills and questions into their representations. To the best of our knowledge, our method GIKT is the first work to model question-skill relations via graph neural network.

\section{Preliminarilies}
\subsubsection{Knowledge Tracing.} In the knowledge tracing task, students sequentially answer a series of questions that the online learning platforms provide. After the students answer each question, 
a feedback of whether the answer is correct will be issued. Here we denote an exercise as $\bm{x}_i=(q_i,a_i)$, where $q_i$ is the question ID and $a_i\in\left\{0,1\right\}$ represents whether the student answered $q_i$ correctly. Given an exercise sequence $\pmb{X} =\left\{\bm{x}_1,\bm{x}_2,...,\bm{x}_{t-1}\right\}$ and the new question $q_t$, the goal of KT is to predict the probability of the student correctly answering it $p(a_t=1|\pmb{X},q_t)$.
\subsubsection{Question-Skill Relation Graph.} Each question $q_i$ corresponds to one or more skills $\left\{s_1,...,s_{n_i}\right\}$, and one skill $s_j$ is usually related to many questions $\left\{q_1,...,q_{n_j}\right\}$, where $n_i$ and $n_j$ 	are the number of skills related to question $q_i$ and the number of questions related to skill $s_j$ respectively. Here we denote the relations as a question-skill relation bipartite graph $\mathcal{G}$, which is defined as $\left\{(q,r_{qs},s)|q\in\mathcal{Q},s\in\mathcal{S}\right\}$, where $\mathcal{Q}$ and $\mathcal{S}$ correpsond to the question and skill sets respectively. 
And $r_{qs}=1$ if the question $q$ is related to the skill $s$. 

\section{The Proposed Method GIKT}
\begin{figure}
	\includegraphics[width=\textwidth]{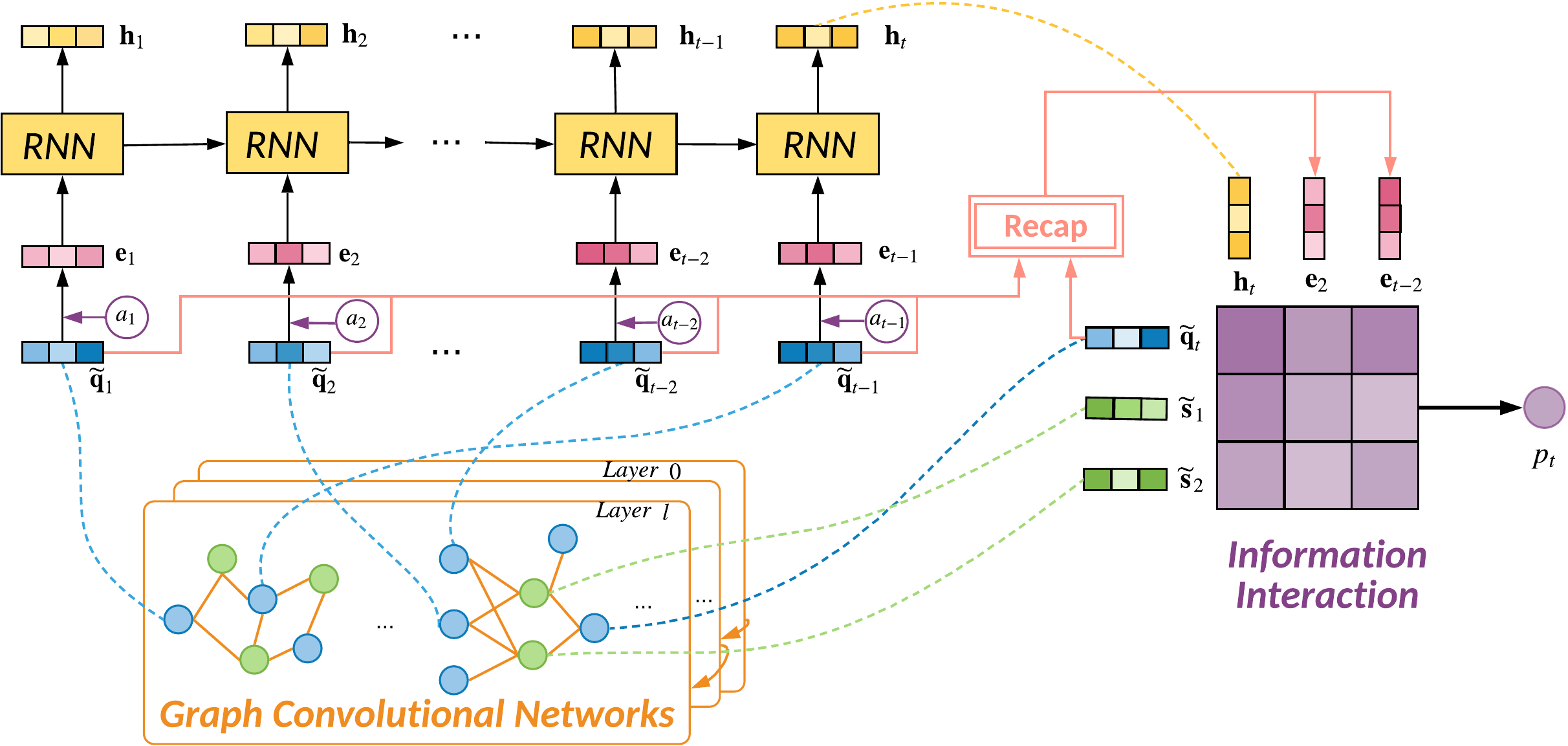}
	\caption{An illustration of GIKT at time step $t$, where $q_t$ is the new question. First we use GCN to aggregate question and skill embeddings. Then a recurrent neural network is used to model the sequential knowledge state $\mathbf{h}_t$. In recap module we select the most related hidden exercises of $q_t$, which corresponds to soft selection and hard selection implementation. The information interaction module performs pairwise interaction between the student’s current state, the selected student’s history 
		exercises, the target question and related skills for the prediction $p_t$.} \label{framework}
	\label{fig:framework}
\end{figure}

In this section, we will introduce our method in detail, and the overall framework is shown in Figure \ref{fig:framework}. We first leverage GCN to learn question and skill representations aggregated on the question-skill relation graph, and a recurrent layer is used to model the sequential change of knowledge state. To capture long term dependency and exploit useful information comprehensively, we then design a recap module followed by an interaction module for the final prediction. 

\subsection{Embedding Layer}

Our GIKT method uses embeddings to represent questions, skills and answers. Three embedding matrices $\mathbf{E}_s\in\mathbb{R}^{|\mathcal{S}|\times d}$, $\mathbf{E}_q\in\mathbb{R}^{|\mathcal{Q}|\times d}$,
$\mathbf{E}_a\in\mathbb{R}^{2 \times d}$ are denoted for look-up operation where $d$ stands for the embedding size. Each row in $\mathbf{E}_s$ or $\mathbf{E}_q$  corresponds to a skill or a question. The two rows in $\mathbf{E}_a$ represent incorrect and correct answers respectively. For $i$-th row vector in matrices, we use $\mathbf{s}_i$, $\mathbf{q}_i$ and $\mathbf{a}_i$ to represent them respectively.

In our framework, we do not pretrain these embeddings and they are trained by optimizing the final objective in an end-to-end manner.

\subsection{Embedding Propagation}


From training perspective, sparsity in question data raises a big challenge to learn informative question representations, especially for those with quite limited training examples. From the inference perspective, whether a student can answer a new question correctly depends on the mastery of its related skills and the question characteristic. When he/she has solved similar questions before, he/she is more likely answer the new question correctly. In this model, we incorporate question-skill relation graph $\mathcal{G}$ to solve sparsity, as well as to utilize prior correlations to obtain better question representations.


Considering the question-skill relation graph is bipartite, the 1st hop neighbors of a question should be its corresponding skills, and the 2nd hop neighbors should be other questions sharing same skills. To extract the high-order information, we leverage graph convolutional network (GCN)  \cite{kipf2016semi} to encode relevant skills and questions into question embeddings and skill embeddings.


Graph convolutional network stacks several graph convolution layers to encode high-order neighbor information, and in each layer the node representations can be updated by embeddings of itself and neighbor nodes. Denote the representation of node $i$ in the graph as $\mathbf{x}_i$ ($\mathbf{x}_i$ can represent skill embedding $\mathbf{s}_i$ or question embedding $\mathbf{q}_i$) and the set of its neighbor nodes as $\mathcal{N}_i$, then the formula of $l$-th GCN layer can be expressed as:

\begin{equation}
\mathbf{x}^l_{i} = \sigma(\frac{1}{|\mathcal{N}_i|} \sum_{j\in \mathcal{N}_i\cup\left\{i\right\}} \mathbf{w}^l\mathbf{x}^{l-1}_j+\mathbf{b}^l),  
\end{equation}

where $\mathbf{w}^l$ and $\mathbf{b}^l$ are the aggregate weight and bias to be learned in $l$-th GCN layer, $\sigma$ is the non-linear transformation such as ReLU.






After embedding  propagation by GCN, we get the aggregated embedding of questions and skills. We use $\widetilde{\mathbf{q}}$ and $\widetilde{\mathbf{s}}$ to represent the question and skill representation after embedding propagation. For easy implementation and better parallelization, we sample a fixed number of question neighbors (i.e., $n_q$) and skill neighbors (i.e., $n_s$) for each batch. And during inference, we run each example multiple times (sampling different neighbors) and average the model outputs to obtain stable prediction results.

\subsection{Student State Evolution}


For each history time $t$, we concatenate the question and answer embeddings and project to $d$-dimension through a non-linear transformation as exercise representations:
\begin{equation}
\mathbf{e}_t = \text{ReLU}(\mathbf{W}_1([\widetilde{\mathbf{q}}_t,\mathbf{a}_t])+\mathbf{b}_1),
\end{equation}
where we use $[,]$ to denote vector concatenation.

There may exist dependency between different exercises, thus we need to model the whole exericise process to capture the student state changes and to learn the potential relation between exercises. To model the sequential behavior of a student doing exercise, we use LSTM \cite{hochreiter1997long} to learn student states from input exercise representations:
\begin{align}
\mathbf{i}_t & = \sigma(\mathbf{W}_i [\mathbf{e}_{t}, \mathbf{h}_{t-1}, \mathbf{c}_{t-1}] + \mathbf{b}_i), \\
\mathbf{f}_t & = \sigma(\mathbf{W}_f [\mathbf{e}_{t}, \mathbf{h}_{t-1}, \mathbf{c}_{t-1}] + \mathbf{b}_f), \\
\mathbf{o}_t & = \sigma(\mathbf{W}_o [\mathbf{e}_{t}, \mathbf{h}_{t-1}, \mathbf{c}_{t-1}] + \mathbf{b}_o), \\
\mathbf{c}_t &= \mathbf{f}_t \mathbf{c}_{t-1}+\mathbf{i}_t \tanh\left(\mathbf{W}_{c} [\mathbf{e}_{t},\mathbf{h}_{t-1}]+\mathbf{b}_c\right),\\
\mathbf{h}_{t} &=\mathbf{o}_t\tanh{\left(\mathbf{c}_t\right)},
\end{align}
where $\mathbf{h}_{t}$, $\mathbf{c}_{t}$,  $\mathbf{i}_t$, $\mathbf{f}_t$, $\mathbf{o}_t$ represents hidden state, cell state, input gate, forget gate, output gate respectively. It is worth mentioning that this layer is important for capturing coarse-grained dependency like potential relations between skills, so we just learn a hidden state $\mathbf{h}_{t}\in\mathbb{R}^d$ as the current student state, which contains coarse-grained mastery state of skills.

\subsection{History Recap Module}

In a student's exercise history, questions of relevant skills are very likely scattered in the long history. From another point, consecutive exercises may not follow a coherent topic. These phenomena raise challenges for LSTM sequence modeling in traditional KT methods: (i) As is well recognized, LSTM can hardly capture long-term dependencies in very long sequences, which means the current student state $\mathbf{h}_t$ may ``forget'' history exercises related to the new target question $q_t$. (ii) The current student state $\mathbf{h}_t$ considers more about recent exercises, which may contain noisy information for the new target question $q_t$. 
When a student is answering a new question, he/she may quickly recall similar questions he/she has done before to help him/her to understand the new question. Inspired from this behavior, we propose to select relevant history exercises(question-answer pair)\footnote{We try other implementations like using history states instead of history exercises , and the results show using history exercises results in a better performance 
	as history states contain other irrelevant information.} $\{\mathbf{e}_i | i \in [1, \dots, t-1]\}$ to better represent a student's ability on a specific question $q_t$, called history recap module. 


We develop two methods to find relevant history exercises. The first one is hard selection, i.e., we only consider the exercises sharing same skills with the new question:
\begin{equation}
\mathcal{\mathbf{I}}_e = \left\{\mathbf{e}_i| \mathcal{N}_{q_i}=\mathcal{N}_{q_t}, i\in [1,..,t-1]\right\},
\end{equation}	

Another method is soft selection, i.e., we learn the relevance between target question and history states through an attention network, and choose top-$k$ states with highest attention scores:
\begin{equation}
\mathcal{\mathbf{I}}_e = \left\{\mathbf{e}_i|R_{i,t}\leq k , V_{i,t}\geq v ,i\in [1,..,t-1]\right\},
\end{equation}
where $R_{i,t}$ is the ranking of attention function $f(\mathbf{q}_i,\mathbf{q}_t)$ like cosine similarity, $V_{i,t}$ is the attention value and $v$ is the lower similarity bound to filter less relevant exercises. 




\subsection{Generalized Interaction Module}

Previous KT methods predict a student's performace mainly according to the interaction between student state $\mathbf{h}_t$ and question representation $\mathbf{q}_t$, i.e., $\langle \mathbf{h}_t, \mathbf{q}_t \rangle$. We generalize the interaction in the following aspects: (i) we use $\langle \mathbf{h}_t, \widetilde{\mathbf{q}}_t \rangle$ to represent the student's mastery degree of question $q_t$, $\langle \mathbf{h}_t, \widetilde{\mathbf{s}}_j \rangle$ to represent the student's mastery degree of the corresponding skill $s_j \in \mathcal{N}_{q_t}$, (ii) we generalize the interaction on current student state to history exercises, which 
reflect the relevant history mastery i.e., $\langle \mathbf{e}_{i}, \widetilde{\mathbf{q}}_t \rangle$ and $\langle \mathbf{e}_i, \widetilde{\mathbf{s}}_j \rangle$, $\mathbf{e}_i \in \mathcal{I}_e$, which is equivalent to let the student to answer the target question in history timesteps.

Then we consider all above interactions for prediction, and define the generalized interaction module. In order to encourage relevant interactions and reduce noise, we use an attention network to learn bi-attention weights for all interaction terms, and compute the weighted sum as the prediction:
\begin{align}
\alpha_{i,j} & = \textup{Softmax}_{i,j}(\pmb{W}^T[\pmb{f}_i,\pmb{f}_j]+b) \\
p_t & = \sum_{\pmb{f}_i\in\mathcal{\mathbf{I}}_e\cup\left\{\mathbf{h}_{t}\right\}}\sum_{\pmb{f}_j\in \mathcal{\widetilde{\mathbf{N}}}_{q_t}\cup\left\{\widetilde{\mathbf{q}}_{t}\right\}} \alpha_{i,j} g(\pmb{f}_i,\pmb{f}_j)
\end{align}

where $p_t$ is the predicted probability of answering the new question correctly, $\mathcal{\widetilde{\mathbf{N}}}_{q_t}$ represents the aggregated neighbor skill embeddings of $q_t$  and we use inner product to implement function $g$. Similar to the selection of neighbors in relation graph, we set a fixed number of $\mathcal{\mathbf{I}}_e$ and $\mathcal{\widetilde{\mathbf{N}}}_{q_t}$ by sampling from these two sets.

\subsection{Optimization}
To optimize our model, we update the parameters in our model using gradient descent by minimizing the cross entropy loss between the predicted probability of answering correctly and the true label of the student's answer:

\begin{equation}
\mathcal{L} = -\sum_t(a_t\log{p_t}+(1-a_t)\log{\left(1-p_t\right)}).
\end{equation}


\section{Experiments}
In this section, we conduct several experiments to investigate the performance of our model. We first evaluate the prediction error by comparing our model with other baselines on three public datasets. Then we make ablation studies on the GCN and the interaction module of GIKT to show their effectiveness in Section \ref{sec:ablation}. Finally, we evaluate the design decisions of the recap module to investigate which design performs better in Section \ref{sec:recap-design}.

\begin{table}
	\centering
	\caption{Dataset statistics}
	\vspace{-5pt}
	\label{tab:stat}
	\begin{tabular}{ccccc}
		\hline
		& ASSIST09 & ASSIST12 & EdNet  \\
		\hline
		\#students & 3,852 & 27,485 & 5000  \\
		\#questions & 17,737 & 53,065 & 12,161  \\
		\#skills & 123 & 265 & 189  \\
		\#exercises & 282,619 & 2,709,436 & 676,974  \\
		questions per skill & 173 & 200 & 147   \\
		skills per question & 1.197 & 1.000 & 2.280  \\
		attempts per question & 16 & 51 & 56  \\
		attempts per skill & 2,743 & 10,224 & 8,420 \\
		\hline
	\end{tabular}
	\label{tab:dataset}
\end{table}

\subsection{Datasets}

To evaluate our model, the experiments are conducted on three widely-used datasets in KT and the detailed statistics are shown in Table \ref{tab:dataset}.
\begin{itemize}
	\item \textbf{ASSIST09}\footnote{\url{https://sites.google.com/site/assistmentsdata/home/assistment-2009-2010-data/skill-builder-data-2009-2010}} was collected during the school year 2009-2010 from ASSISTments online education  platform\footnote{\url{https://new.assistments.org/}}. We conduct our experiments on ``skill-builder" dataset. Following the previous work \cite{xiong2016going}, we remove the duplicated records and scaffolding problems from the original dataset. This dataset has 3852 students with 123 skills, 17,737 questions and 282,619 exercises. 
	
	\item \textbf{ASSIST12}\footnote{\url{https://sites.google.com/site/assistmentsdata/home/2012-13-school-data-with-affect}} was collected from the same platform as ASSIST09 during the school year 2012-2013. In this dataset, each question is only related to one skill, but one skill still corresponds to several questions. After the same data processing as ASSIST09, it has 2,709,436 exercises with 27,485 students, 265 skills and 53,065 questions.
	
	\item \textbf{EdNet}\footnote{\url{https://github.com/riiid/ednet}} was collected by \cite{choi2019ednet}. As the whole dataset is too large, we randomly select 5000 students with 189 skills, 12,161 questions and 676,974 exercises.
	
	

\end{itemize}

Note that for each dataset we only use the sequences of which the length is longer than 3 in the experiments as the too short sequences are meaningless. For each dataset, we split $80\%$ of all the sequences as the training set, $20\%$ as the test set. To evaluate the results on these datasets, we use the area under the curve (AUC) as the evaluation metric.

\subsection{Baselines}
In order to evaluate the effeciveness of our proposed model, we use the following models as our baselines:

\begin{itemize}
	\item \textbf{BKT} \cite{corbett1994knowledge} uses Bayesian inference for prediction, which models the knowledge state of the skill as a binary variable. 
	
	\item \textbf{KTM} \cite{vie2019knowledge} is the latest factor analysis model that uses Factorization Machine to interact each feature for prediction. Although KTM can use many types of feature, for fairness we only use question ID, skill ID and answer as its side information in comparison.

	\item \textbf{DKT} \cite{dkt} is the first method that uses deep learning to model knowldge tracing task. It uses recurrent neural network to model knowldge state of students.
	
	\item \textbf{DKVMN} \cite{dkvmn} uses memory network to store knowledge state of different concepts respectively instead of using a single hidden state.
	
	\item \textbf{DKT-Q} is a variant of DKT that we change the input of DKT from skills to questions so that the DKT model directly uses question information for prediction.
	
	\item \textbf{DKT-QS} is a variant of DKT that we change the input of DKT to the concatenation of questions and skills so that the DKT model uses question and skill information simultaneously for prediction.
	
	\item \textbf{GAKT} is a variant of the model  Exercise-Enhanced Recurrent Neural Network with Attention mechanism (EERNNA) \cite{su2018exercise} as EERNNA utilizes question text descriptions but we can't acquire this information from public datasets. Thus we utilize our input question embeddings aggregated by GCN as input of EERNNA and follow its framework design for comparison.
	
\end{itemize}

\subsection{Implementation Details}
We implement all the compaired methods with TensorFlow. The code for our method is available online\footnote{\url{https://github.com/Rimoku/GIKT} }.
The embedding size of skills, questions and answers are fixed to 100, all embedding matrices are randomly initialized and updated in the training process. In the implementation of LSTM, a stacked LSTM with two hidden layers is used, where the sizes of the memory cells are set to 200 and 100 respectively. In embedding propagation module, we set the maximal aggregate layer number $l = 3$. We also use dropout with the keep probability of $0.8$ to avoid overfitting. All trainable parameters are optimized by Adam algorithm\cite{kingma2014adam} with learning rate of 0.001 and the mini-batch size is set to 32. Other hyper-parameters are chosen by grid search, including the number of question neighbors in GCN, skill neighbors in GCN, related exercises and skills related to the new question.

\subsection{Overall Performance}
Table \ref{tab:performance} reports the AUC results of all the compared methods. From the results we observe that our GIKT model achieves the highest performance over three datasets, which verifies the effectiveness of our model. To be specific, our proposed model GIKT achieves at least $1\%$ higher results than other baselines. Among the baseline models, traditional machine learning models like BKT and KTM perform worse than deep learning models, which shows the effectiveness of deep learning methods. DKVMN performs slightly worse than DKT on average as building states for each concept may lose the relation information between concepts. Besides, GAKT performs worse than our model, which indicates that exploiting high-order skill-question relations through selecting the most related exercises and performing interaction makes a difference.




\begin{table}[t]
	\centering
	\caption{The AUC results over three datasets. Among these models, BKT, DKT and DKVMN predict for skills, other models predict for questions. Note that ``*'' indicates that the statistically significant improvements over the best baseline, with p-value smaller than $10^{-5}$ in two-sided t-test.}
	\begin{tabular}{cccc}
		\hline
		Model & ASSIST09 & ASSIST12 & EdNet\\
		\hline
		BKT & 0.6571 & 0.6204 & 0.6027 \\
		KTM & 0.7169 & 0.6788 & 0.6888 \\
		DKVMN & 0.7550 & 0.7283 & 0.6967  \\
		DKT & 0.7561 & 0.7286 & 0.6822 \\
		
		\hline
		DKT-Q & 0.7328 & 0.7621  &  0.7285 \\
		DKT-QS & 0.7715 & 0.7582 & 0.7428 \\
		GAKT & 0.7684 & 0.7652 & 0.7281\\
		\hline
		GIKT & \textbf{0.7896}* & \textbf{0.7754}* & \textbf{0.7523}*   \\
		
		\hline
	\end{tabular}
	\label{tab:performance}
\end{table}

On the other hand, we find that directly using questions as input may achieve superior performance than using skills. For the question-level model DKT-Q, it has comparable or better performance than DKT over ASSIST12 and EdNet datasets. However, DKT-Q performs worse than DKT in ASSIST09 dataset. The reason may be that the average number of attempts per question in ASSIST09 dataset is significantly less than other two datasets as observed in Table \ref{tab:stat}, which illustrates DKT-Q suffers from data sparsity problem. 
Besides, the AUC results of the model DKT-QS are higher than DKT-Q and DKT, except on ASSIST12 as it is a single-skill dataset, which indicates that considering question and skill information together improves overall performance. 




\subsection{Ablation Studies}
\label{sec:ablation}
To get deep insights on the effect of each module in GIKT, we design several ablation studies to further investigate on our model. We first study the influence of the number of aggregate layers, and then we design some variants of the interaction module to investigate their effectiveness.


\subsubsection{Effect of Embedding Propagation Layer}	
We change the number of the aggregate layers in GCN ranging from 0 to 3 to show the effect of the high-order question-skill relations and the results are shown in Table \ref{tab:layers}. Specially, when the number of the layer is 0, it means the question embeddings and skill embeddings used in our model are indexed from embedding matrices directly. 



\begin{table}
	\centering
	\caption{Effect of the number of aggregate layers.}
	\begin{tabular}{cccc}
		\hline
		Layers & ASSIST09 & ASSIST12 & EdNet  \\
		\hline
		
		\hline
		0 & 0.7843 & 0.7738 & 0.7438    \\
		1 & 0.7844 & 0.7710 & 0.7432   \\
		2 & 0.7894 & 0.7736 & 0.7466  \\
		3 & \textbf{0.7896} & \textbf{0.7754} & \textbf{0.7523}  \\

		\hline
	\end{tabular}
	\label{tab:layers}
\end{table}

From Table \ref{tab:layers} we find that, when the number of aggregate layer from zero to one, the performance of GIKT changes slightly, as we have already used 1-order relation in recap module and interaction module. However, GIKT achieves better performance when the number of aggregate layers increases, which validates the effectiveness of GCN. The results also imply that exploiting high-order relations contained in the question-skill graph is necessary for adequate results as the performance of adopting more layers is better than using less layers.

\subsubsection{Effect of Interaction Module}

To verify the impact of interaction module in GIKT, we conduct ablation studies on four variants of our model. The details of the four settings are listed as below and the performance of them is shown in Table \ref{tab:info}. 
\begin{itemize}

	\item \textbf{GIKT-RHS} (Remove History related exercises and Skills related to the new question) For GIKT-RHS, we just use the current state of the student and the new question to perform interaction for prediction.
	\item \textbf{GIKT-RH} (Remove History related exercises) For GIKT-RH, we only use the current state of the student to model mastery of the new question and related skills.
	\item \textbf{GIKT-RE} (Remove Skills related to the new question)  For GIKT-RS, we do not model the mastery of skills related to the new question.
	\item \textbf{GIKT-RA} (Remove Attention in interaction module)  GIKT-RA removes the attention mechanism after interaction, which treats each interaction pair as equally important and average the prediction scores directly in interaction part for prediction.
\end{itemize}

\begin{table}
	\centering
	\caption{Effect of Information Module}
	\begin{tabular}{cccc}
		\hline
		Model & ASSIST09 & ASSIST12 & EdNet \\
		\hline
		
		\hline
		GIKT-RHS & 0.7814 & 0.7672 &  0.7420  \\
		GIKT-RH & 0.7808 & 0.7703  &  0.7463 \\
		GIKT-RS & 0.7864  & 0.7754 & 0.7428  \\
		GIKT-RA & 0.7856  & 0.7711 & 0.7500\\
		GIKT & \textbf{0.7896}  & \textbf{0.7754} &  \textbf{0.7523} \\
		\hline
	\end{tabular}
	\label{tab:info}
\end{table}

From Table \ref{tab:info} we have the following findings: Our GIKT model considering all interaction aspects achieve best performance, which shows the effectiveness of the interaction module. Meanwhile, from the results of GIKT-RH we can find that relevant history states can help better model the student's ability on the new question. Besides, the performance of GIKT-RS is slightly worse than GIKT, which implies that model mastery degree of question and skills simultaneously can further help prediction. Note that as ASSIST12 is a single-skill dataset, use skill information in interaction module is redundant after selecting history exercises sharing the same skill, thus we set the number of sampled related skill as 0. Comparing the results of GIKT-RA with GIKT, the worse performance confirms the effectiveness of attention in interaction module, which distinguishes different interaction terms for better prediction results. By calculating different aspects of interaction and weighted sum for the prediction, information from different level can be fully interacted. 





\subsection{Recap Module Design Evaluation}
\label{sec:recap-design}


To evaluate the detailed design of the recap module in GIKT, we conduct experiments of several variants. The details of the settings are listed as below and the performance of them is shown in Table \ref{tab:interaction}.
\begin{itemize}
	
	
	
	\item \textbf{GIKT-HE} (Hard select history Exercises) For GIKT-HE, we select the related exercises sharing the same skills.
	
	\item \textbf{GIKT-SE} (Soft select history Exercises) For GIKT-SE, we select history exercises according to the attention weight.
	
	\item \textbf{GIKT-HS} (Hard select hidden States) For GIKT-HS, we select the related hidden states of the exercises sharing the same skills.
	
	\item \textbf{GIKT-SS} (Soft select hidden States) For GIKT-SS, we select hidden states according to the attention weight. The reported results of GIKT in previous sections are taken by the performance of GIKT-SS.
	
\end{itemize}

\begin{table}
	\centering
	\caption{Results of Different Recap Module Design}
	\begin{tabular}{ccccc}
		\hline
		Model & ASSIST09 & ASSIST12 & EdNet \\
		\hline
		
		GIKT-HE & \textbf{0.7896} & \textbf{0.7753} & 0.7481  \\
		GIKT-SE & 0.7870 & 0.7686 &  \textbf{0.7523}  \\
		GIKT-HS & 0.7788 & 0.7672 & 0.7364\\
		GIKT-SS & 0.7743 & 0.7683 & 0.7417  \\
		\hline
		\hline
	\end{tabular}
	\label{tab:interaction}
\end{table}

From Table \ref{tab:interaction} we find that selecting history exercises performs better than selecting hidden states. This result implies that the hidden state contain irrelevant information for the next question as it learns a general mastery for a student. Instead, selecting exercises directly can reduce noise to help prediction. The performances of hard selection and soft selection distinguish on different datasets. Using attention mechanism can achieve better selection coverage while the hard selection variant can select exercises via explicit constraints.

\section{Conclusion}
In this paper, we propose a framework to employ high-order question-skill relation graphs into question and skill representations for knowledge tracing. Besides, to model the student's mastery for the question and related skills, we design a recap module to select relevant history states to represent student's ability. Then we extend a generalized interaction module to represent the student's mastery degree of the new question and related skills in a consistent way. To distinguish relevant interactions, we use an attention mechanism for the prediction. The experimental results show that our model achieve better performance.

\section*{Addendum Version}
After the deadline of submitting the camera-ready version to the proceedings, we found our realization of ``soft selection" mentioned in Section 4.4 is somewhat unreasonable, thus we adopt another suitable realization for this strategy, which causes some differences of the results with the proceeding version. This version is the newest version and we report the revised experiment results in this version with the code. As the conference proceedings have been prepared already, PC chairs suggested us to upload an addendum version by ourself. Please refer to this version.

%
%
%

	%
	%
	%
	
\end{document}